\documentclass[letterpaper, 10 pt, conference]{ieeeconf}  

\IEEEoverridecommandlockouts                              

\usepackage{graphicx}
\usepackage{times} 
\usepackage{amsmath} 
\usepackage{amssymb}  
\usepackage{cite}
\usepackage[english]{babel}
\usepackage{balance}
\usepackage{xcolor}

\usepackage{float}
\usepackage{colortbl}
\usepackage{booktabs} 

\usepackage{geometry}
\usepackage{algorithm}
\usepackage[noend]{algpseudocode}

\usepackage{caption}
\title{\LARGE \bf
Contact-Aided Navigation of Flexible Robotic Endoscope Using Deep Reinforcement Learning in Dynamic Stomach
}

\author{
Chi Kit Ng, Huxin Gao, Tian-Ao Ren, Jiewen Lai, Hongliang Ren%
\thanks{This work was supported by the Hong Kong Research Grants Council Collaborative Research Fund under Grant CRF-C4026-21G, by the Hong Kong Research Grants Council Research Impact Fund under Grant RIF-R4020-22, Hong Kong RGC GRF (14204524, 14203323, 14216022), NSFC/RGC Joint Research Scheme N\_CUHK420/22.}%
\thanks{C. Ng, H. Gao, J. Lai, and H. Ren are with the Department of Electronic Engineering, The Chinese University of Hong Kong, Shatin, N.T., Hong Kong, China. {\tt\small hlren@ee.cuhk.edu.hk}}%
\thanks{T.-A. Ren is with the Department of Mechanical Engineering, Stanford University, Stanford, CA 94305, USA.}%
}
\begin{document}

\maketitle

\begin{abstract}
 Navigating a flexible robotic endoscope (FRE) through the gastrointestinal tract is critical for surgical diagnosis and treatment. However, navigation in the dynamic stomach is particularly challenging because the FRE must learn to effectively use contact with the deformable stomach walls to reach target locations. To address this, we introduce a deep reinforcement learning (DRL) based Contact-Aided Navigation (CAN) strategy for FREs, leveraging contact force feedback to enhance motion stability and navigation precision. The training environment is established using a physics-based finite element method (FEM) simulation of a deformable stomach. Trained with the Proximal Policy Optimization (PPO) algorithm, our approach achieves high navigation success rates (within 3 mm error between the FRE’s end-effector and target) and significantly outperforms baseline policies. In both static and dynamic stomach environments, the CAN agent achieved a 100\% success rate with 
 1.6 mm average error, and it maintained an 85\% success rate in challenging unseen scenarios with stronger external disturbances. These results validate that the DRL-based CAN strategy substantially enhances FRE navigation performance over prior methods.
\end{abstract}


\section{Introduction}
\begin{figure}[t!]
  \centering
  \includegraphics[scale=0.08]{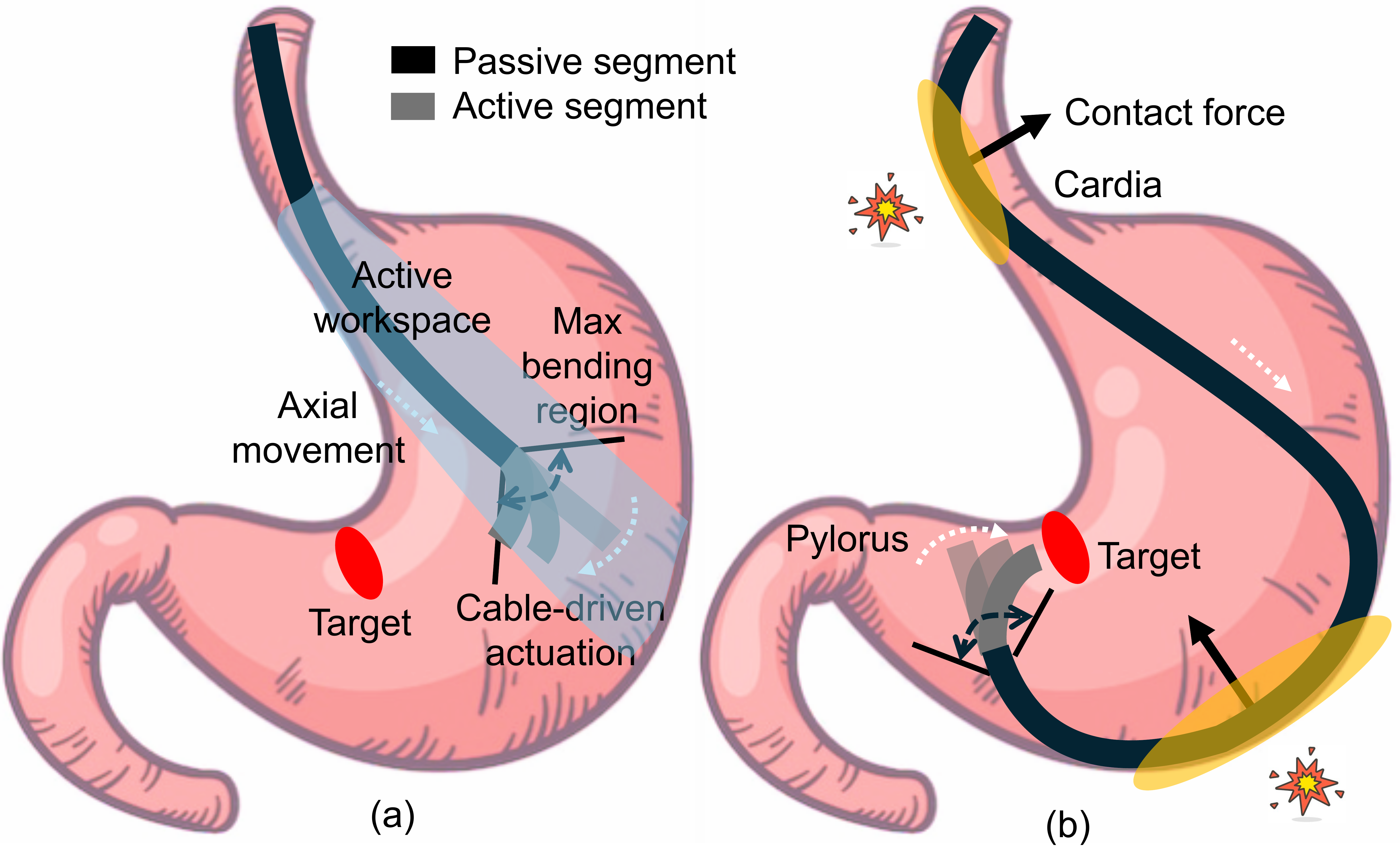}
  \caption{Description of CAN strategy. (a) Without interaction with the stomach wall, actuations (translation, rotation and bending) can barely navigate the tip of the endoscope to the target. (b) Leveraging contact between endoscope and stomach wall to navigate towards the target.}
  \label{Fig:introfig}
\end{figure}

Over the past decade, gastrointestinal endoscopy has become a cornerstone for both surgical diagnostics and treatment \cite{chiu2024EndoMaster, gaoijrr}. Traditional manual endoscopy faces significant challenges in accessing certain regions within the stomach due to the tissue deformation, the dynamic movement of that stomach and non-intuitive manual manipulation. To tackle this, an increasing number of flexible robotic endoscopes (FRE) were developed \cite{laitii,chiu2024EndoMaster, gaoijrr}.
The FRE, a type of continuum robot, offers inherent compliance, enabling it to adapt its shape to tortuous gastrointestinal (GI) tracts \cite{crreviewTRO}. The active segment has three degrees of freedom (DoFs) of our FRE, which include two bendings and one translational movement. As shown in Fig. \ref{Fig:introfig}(a), the FRE consists of 1000 mm passive segments and a 79 mm active segment (7\% of entire endoscope), with a bending range over $\pm 90^{\circ}$ \cite{gaoijrr}. The average length of the male stomach is 190 mm, and the average length of the greater curvature of the male stomach is 344 mm \cite{stosize}. Without utilizing contact with the stomach walls, the reachable workspace of the endoscope tip is greatly limited. As illustrated in Fig. \ref{Fig:introfig}(a), purely executing tip bending, base rotation, and insertion motions (without wall contact) cannot guide the endoscope tip to certain targets on the stomach wall. The passive segment simply floats in the large gastric cavity, and the tip cannot reach deep targets by free-space motions alone. In contrast, if the endoscope can brace or push against the stomach walls (Fig. \ref{Fig:introfig}(b)), the environment contact provides a form of “external constraint” or fulcrum that the robot can leverage to extend its reach and stability. In other words, contact is indispensable for navigating FREs inside expansive and deformable environments like the stomach. This insight underpins the concept of Contact-Aided Navigation (CAN): using the endoscope’s contact with the surroundings (here, the stomach wall) to help guide the robot towards the target.

Several research efforts have recognized the importance of contact in continuum robot navigation. For instance, Zhang et al. implemented a motion controller for a tendon-driven continuum catheter that explicitly modeled contacts using FEM, enabling real-time contact force estimation and control without relying on any external sensors \cite{zhang2019motion}. Recent work in motion planning has also embraced contact-rich strategies. Rao et al. developed a contact-aided motion planner that leverages environmental contact to improve path planning for tendon-driven continuum robots \cite{softcontactnav2}. In scenarios where avoiding collisions is impossible, contact-inclusive planners have proven more robust than traditional purely collision-avoiding methods. For example, researchers have shown that a soft “vine” robot can navigate more reliably by exploiting contact with obstacles rather than treating them as forbidden zones \cite{softcontactnav3}. 

Furthermore, new sensing and control frameworks are emerging to handle contacts: Wang et al. proposed a perception-action loop that compares expected vs. actual robot shape to detect contact events and adjust a soft robot’s configuration in real time \cite{nc2024senseexpectation}. Most existing research on CAN focuses on sparse contact situations \cite{softcontactnav3, softcontactnav1, nc2024senseexpectation}, or studies CAN in small, static cavity regions \cite{haggerty2019characterizing, zhang2019motion}. Extending CAN strategies to a large, deformable, and dynamically moving environment like the stomach poses new challenges that have not been fully addressed by existing approaches.

Deep reinforcement learning (DRL) has emerged as a promising tool for enabling FREs to learn CAN strategies in complex, dynamic, contact-rich environments. DRL allows robotic systems to learn sophisticated policies through trial-and-error interactions with the environment \cite{ng2024navigation, softdrlsimreal, softdrlsofagym, softdrldr, (JIRS23)RL_CR_control, nce24Axisspace, elastica}. Unlike traditional path planning methods that rely on predefined models or heuristics, DRL is a data-driven approach that offers the flexibility and adaptability needed to navigate dynamic, uncertain environments. Notably, Ng et al. demonstrated the feasibility of training a tendon-driven flexible endoscope agent using PPO in the SOFA simulation environment \cite{ng2024navigation}. However, this work did not leverage an explicit CAN strategy and was evaluated in a simpler scenario. It lacked the dynamic physiological motions of the stomach and did not integrate force feedback into the policy’s observations. As a result, the learned policy could struggle when collisions occur or when the environment is dynamic. By doing so, our method enables the FRE to intelligently utilize contacts with the stomach wall to navigate toward targets, even as the stomach undergoes deformations and motion.

In this work, we leverage the contact feedback to improve endoscopic navigation capability in deformable and dynamic stomachs. The major contributions are as follows:
\begin{itemize} 
\item Innovative DRL-based CAN strategy: We develop a novel CAN approach, integrating force feedback into the DRL framework. This enables the FRE to leverage contact forces and interaction dynamics, improving adaptability and performance in dynamic stomachs.

\item Dynamic simulation environment: We construct a dynamic and deformable stomach environment in physics-based SOFA simulation that assembles breathing, heartbeat, and subtle movements of the human body during FRE navigation. The training results show the FRE needs real-time adaptation to environmental deformations and external disturbances, which is critical for autonomous endoscopic navigation in clinical scenarios.

\item Generalization and robustness: We conduct extensive zero-shot evaluations in unstructured environments (UE) with unseen target locations and varying external force dynamics, demonstrating the adaptability of policy trained by CAN strategy.

\end{itemize}
\section{Method}
This section shows the dynamic stomach environment, which is constructed using the physics-based SOFA simulator. FEM models the deformable stomach and real-time contact detection accurately captures the interactions between the robot and its surroundings. Then, a DRL training pipeline in constructed environments is illustrated.

\begin{figure}[htbp] \centering \includegraphics[scale=0.13]{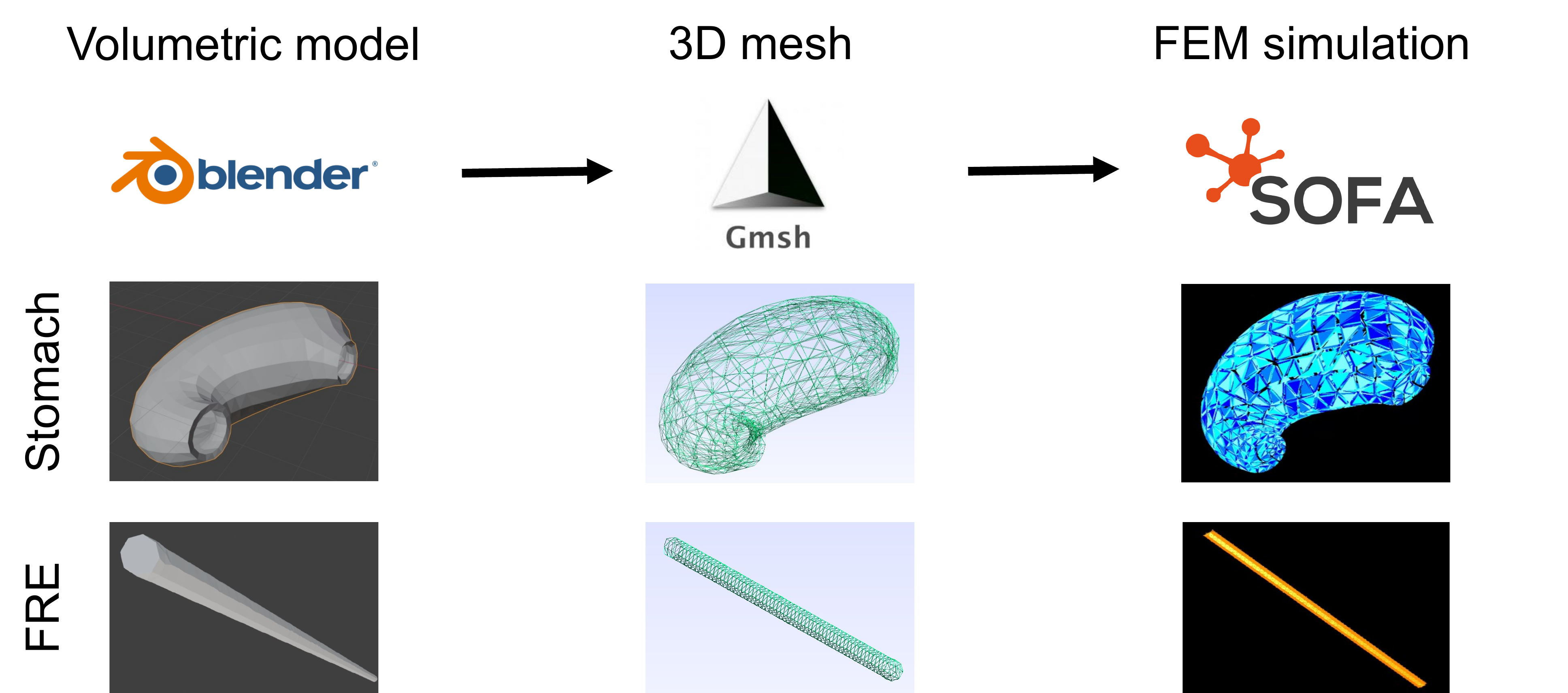} \caption{Volumetric deformable models of the stomach and endoscope were created using Blender (Blender Foundation, Netherlands). The 3D mesh was generated using the open-source software Gmsh \cite{gmsh} and imported into the SOFA physics engine \cite{sofa} as a mechanical and collision model, respectively.} \label{Fig: modeling} 
\end{figure}

\subsection{Force-informed dynamic simulation environment}
\label{method:env}
Efficient contact detection pipelines and model order reduction methods enable real-time deformation and force modeling \cite{sofa}. FEM is widely used for soft robotics simulation due to its ability to represent complex geometries and non-linear material behaviors. A denser mesh can capture finer curvature details. As shown in Fig. \ref{Fig: modeling}, FEM discretizes a curved structure into smaller, manageable elements. To reduce computational complexity, these elements are defined as nodes, dividing the continuous domain of the soft body using a discretization method.
\begin{equation}
u(x) = \sum^n_{i=1}N_i(x)u_i,
\label{Equ:discretization}    
\end{equation}
where $u(x)$ represents the displacement at position x, $N_i(x)$ is shape function and $u_i$ is nodal displacement. External forces, such as pulling forces and gravity, induce deformation in the soft body. In volumetric meshes, tetrahedral elements define the DoFs of the deformable object. Within the simulation, force propagates from the outer regions to the inner regions. To model the deformable behavior of the stomach, a volumetric mesh is generated using Gmsh. The system dynamics are governed by the equation of motion:
\begin{equation}
M\frac{\mathrm{d}^2u}{\mathrm{d}t^2}+C\frac{\mathrm{d}u}{\mathrm{d}t}+K_eu = F_{external},
\label{Equ:dynamics}    
\end{equation}
where $M$ is the mass matrix, $u$ is the displacement vector as defined in Eq. \ref{Equ:discretization}, $C$ is the damping matrix, $K_e$ is the stiffness matrix to specify material properties such as Young’s modulus and Poisson’s ratio, and $F_{external}$ represents external forces. Internal forces propagate within the volumetric tetrahedral mesh and are modeled as $f_{int}=K_e u$. These forces are used to compute the displacement of each DoF at each time step, ensuring accurate force propagation and system dynamics.

\begin{figure}[t!] \centering \includegraphics[scale=0.3]{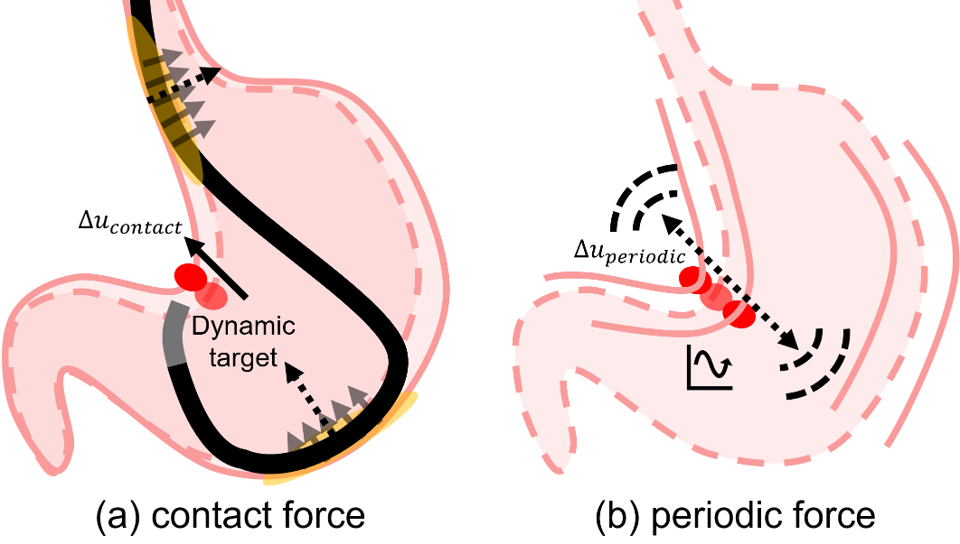} \caption{Dynamic environment modeling: (a) Contact forces generated by the interaction between the robotic endoscope and the deformable stomach. (b) Applied sinusoidal forces simulating periodic gastric motility and breathing-induced movements.} \label{Fig:dynamiccases} \end{figure}

As shown in Fig.~\ref{Fig:dynamiccases}(a), the FRE interacts with a deformable stomach, which introduces displacement $\Delta u_{contact}$ of the target point due to contact force. Real-time contact handling in FEM ensures that two soft bodies do not interpenetrate and efficiently manage collisions. Contact and collision generate reaction forces that push the contact points away from the surface. A constraint-based method is employed to maintain zero penetration by defining a constraint function $g(u)$ between two objects $u_A$ and $u_B$.  Given a gap p between them, the constraint function is formulated as
\begin{equation}
g(u)=p, \quad u={u_A,u_B}.
\label{Equ: contactconstraint }    
\end{equation}
The Lagrange multiplier method is added into the $F_{external}$ of Eq. \ref{Equ:dynamics} to handle contact forces $F_c$, which is formulated by 
\begin{equation}
F_c = G^T \lambda,
\label{Equ:contacthandling }    
\end{equation}
where $G$ is matrix of contact constraints, and $\lambda$ is the Lagrange multipliers.

\begin{algorithm}
\begin{scriptsize} 
\caption{ESTABLISH SIMULATION ENVIRONMENT}
\label{Algo1}
\begin{algorithmic}[1]
\Procedure{ROOTNODE}{}
    \State \textbf{requiredPlugins} \Comment{Add plugins to MainHeader}
    \State \textbf{defaultVisualManagerLoop} \& \textbf{freeMotionAnimationLoop}
    \State \textbf{visualStyle} \& \textbf{gravity} \Comment{$G = 9.8 \, \mathrm{m/s^2}$}
    \State \textbf{collisionPipeline}: alarmDistance=2, contactDistance=0.5, frictionCoef=0.1

    \State \textbf{DynamicEnvNode()}
    \State \hspace{1em} \textbf{DeformablePhantomModel}: FEM, visual, collisionModel \Comment{\texttt{vtk, obj}}
    \State \hspace{2em} \textbf{FixedLagrangianConstraint} \Comment{indices {85 72 107}} 
    \State \hspace{2em} \textbf{LinearForceAlternator()}:  \Comment{indices {84 267}, force=1N}

    \State \textbf{SoftRobotNode()}
    \State \hspace{1em} \textbf{softRobot}: FEM, visual, collisionModel \Comment{\texttt{vtk, obj}}
    \State \hspace{2em} \textbf{deformablePart}: cableActuators \Comment{Front end with four cables}
    \State \hspace{2em} \textbf{Controller}: Sofa.Core.Controller
    \State \hspace{3em} \textbf{PolicycontrolNode}: GymStepController() 

    \State \textbf{define frames}: target list \& end-effector

    \State \textbf{ContactListener}: getContactData() \Comment{collisionPointsModel}
\EndProcedure
\end{algorithmic}
\end{scriptsize} 
\end{algorithm}
To facilitate reinforcement learning in a force-informed dynamic environment, Algorithm \ref{Algo1} outlines the simulation setup, which integrates FRE with contact information for training agents. The FRE consists of four cable-driven actuation units and one axial DoF. The environment is designed to simulate endoluminal conditions. The simulation begins with the initialization of the environment (ROOTNODE function in SOFA). Essential plugins are added to the main header to enable required functionalities for solving FEM force field, visualization, collision detection etc. Threshold values for collision detection are set at 2 mm for alarms and 0.5 mm for contact, with a friction coefficient of 0.1 to mimic realistic surface interactions (handled by the collisionPipeline in SOFA).
Dynamic properties of the environment include a deformable phantom model representing the stomach, which is simulated using FEM with collision and visual components (DynamicEnvNode). Specific indices of the volumetric stomach mesh are assigned as fixed constraints. When subjected only to constant forces, the internal forces in the stomach reach an equilibrium state, stabilizing the structure. As shown in Fig.~\ref{Fig:dynamiccases}(b), to introduce dynamic environmental conditions, discrete periodic external forces with a period of $2T$ to introduce environmental dynamics, which is defined as
\begin{equation}
\vec{F_p}(t) = 
\begin{cases} 
\vec{f}_0, & \text{if } \left( \frac{t}{T} \right)\mod 2 \in [0,1), \\ 
-\vec{f}_0, & \text{if } \left( \frac{t}{T} \right)\mod 2 \in [1,2).
\end{cases}
\label{Equ:periodicforce}    
\end{equation}
FRE includes a deformable section actuated by tendon-driven mechanisms, which control the front-end motion of the robot (SoftRobotNode). The robot's operation is managed by a controller within the reinforcement learning framework (GymStepController). 
Real-time contact data is collected using the ContactListener component, which extracts collision information from collisionPointsModel (i.e., collision between stomach and FRE). This ensures that contact force data is continuously monitored and integrated into the simulation pipeline. The resultant force and contact feedback are then used as state inputs for reinforcement learning, allowing the agent to adapt to dynamic endoluminal conditions and optimize navigation strategies.

\subsection{Deep reinforcement learning}
\begin{figure}[htbp]
  \centering
  \includegraphics[scale=0.10]{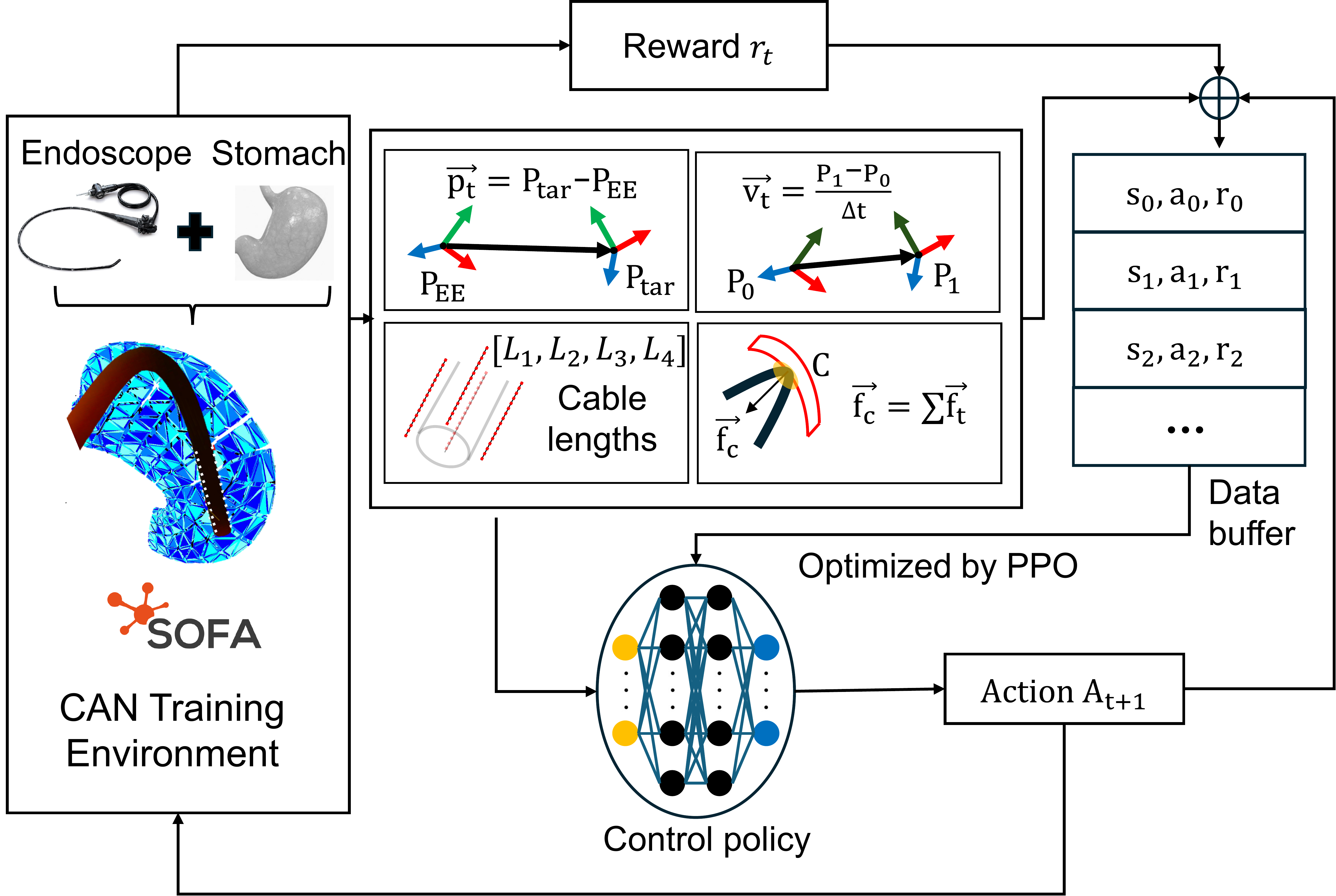}
  \caption{Force-informed deep reinforcement training pipeline in SOFA framework}
  \label{Fig:system architecture}
\end{figure}

On-policy DRL trains an agent to navigate by learning from interactions with its environment. The Markov Decision Process (MDP) provides a framework for modeling sequential decision-making problems in RL. Elements {$S,A,R,P, \gamma$} are defined in the feedback control loop of CAN problems, which can be formulated as a long-horizon discounted MDP. $s_t \in S $ is the state at the time $t$, $a_t \in A$ is the action in time $t$, and the reward function $r_t=R(s_t,a_t)$. The transition dynamics, $P(s_{t+1}|s_t,a_t)$, describe the probability of reaching state $s_{t+1}$ given the current state $s_t$ and $a_t$. The discount factor $\gamma \in [0,1]$ determines the less importance of past rewards, ensuring long-term optimization in the decision-making process.

At the beginning of training, a simulation scene is created using a predefined method described in Algorithm~\ref{Algo1}. To ensure robust policy generalization, the target position is randomized via a reset function, and the initial state of the environment is recorded as the observation vector. As shown in Fig. \ref{Fig:system architecture}, the state $s_t$ consists of $\vec{p_t}, \vec{v_t},L_t,\vec{F_t}, C$, where $\vec{p_t}$ is the position vector between end-effector (EE) and target, $\vec{v_t}$ is the EE's instantaneous velocity, $L_t$ represents four cable lengths, $\vec{F_t}$ is the resultant contact force vector, and $C$ is the contact indicator. The reward function $r_t$ consists of $R_{dis}, R_B, R_S$, defined as:
\begin{align}
    R_{dis} &= -\|P_{EE} - P_{tar}\|_2 \\
    R_B &= 
    \begin{cases} 
        -10^4, & x_{EE} < 15 \\
        0, & \text{otherwise}
    \end{cases} \\
    R_S &= 
    \begin{cases} 
        10^4, & \|P_{EE} - P_{tar}\|_2 < 3 \\
        0, & \text{otherwise}
    \end{cases}
\label{Eq:reward setting3}
\end{align}
$R_{dis}$ specifies the negative reward of the distance between EE and the target point. $R_B$ indicates a boundary penalty when EE leaves the pylorus stomach. $R_S$ is a positive reward in the case of reaching the targets.
\begin{algorithm}
\begin{scriptsize} 
\caption{POLICY TRAINING WITH CONTACT FORCE INFORMATION of STATE-BASED DRL}
\label{Algo2}
\textbf{Input:} 
\begin{itemize}
    \item Kinematic information of End-effector \Comment{position vector to target, velocity}
    \item Actuation values of previous timestep
    \item Contact information \Comment{boolean value for contact or not, resultant force vector}
\end{itemize}
\textbf{Output:} Parameters of control policy Neural Network $\pi$ \\
\textbf{Initialization:} Network parameters
\begin{algorithmic}[1]

\State Create $environment$ with Gymnasium \cite{towers2024gymnasium}
\State $env \gets \text{gym.make(endogym-v0)}$
\For{$t < \text{TOTAL\_TIMESTEPS}$}
\If{$t = 0$}
    \State $node \gets Sofa.Core.Node(``root")$
    \State $ROOTNODE \gets createScene(node)$ \Comment{created by Algorithm 1}
    \State $target \gets resettarget(seed)$
    \State $obs \gets \text{initial observation}$
    
\EndIf
\State \textbf{End if}
\State $action \gets \pi(obs)$
\State $obs \gets env.get\_state(action)$ 
\State $\alpha \gets \text{1  if end-effector is out of bound, else 0} $
\State $\beta \gets \text{1  if end-effector reaches target, else 0} $
\State $reward \gets -\alpha10^4+\beta10^4- \|\text{target} - \text{obs}\|_2$
\If{Episode end}
\State $\pi \text{ updated by } \text{PPO}$ \cite{SB3}
\EndIf

\State $t \gets t + 1$
\EndFor
\State \textbf{End for}
\end{algorithmic}
\textbf{Output:} $\pi$
\end{scriptsize}
\end{algorithm}

The control policy is defined as a deep neural network. Based on neural networks, FRE can predict the dynamics of the system \cite{trainlearn}, where perception is the input, and action is the output. As shown in Algorithm~\ref{Algo2}, the training process begins with the initialization of a simulation environment using Gymnasium \cite{towers2024gymnasium}, specifically the custom environment endogym-v0, which integrates the SOFA simulation framework. This environment allows for real-time interaction between the policy and the simulated FRE. The algorithm iteratively trains an optimal control policy, denoted $\pi$, by mapping the current observation to an action. The action is then applied to the environment, which updates the robot's state. Sequential data pairs are stored in the data buffer. We adopt the Proximal Policy Optimization (PPO) algorithm due to its robust, stable performance in high-dimensional, continuous control tasks \cite{PPO}.

\section{Experiment}
We conducted a series of experiments in the simulation environment to assess the performance of the proposed CAN strategy. The experiments are designed to answer three questions: (1) Does incorporating contact force feedback into the state improve navigation performance? (2) How does training in a dynamic environment (with moving walls) compare to training in a static environment? (3) Can the learned policy generalize to new, unseen scenarios (targets and dynamics) that were not encountered during training?
\subsection{Experiment setup}
\subsubsection{Environment}
The FREs were trained and evaluated in environments with varying conditions, defined as follows:
\begin{itemize}
\item Free environment (FE): an empty environment where the FRE can reach targets without contact.
\item Static environment (SE): ensures deformation only occurs when the robotic endoscope interacts with it. A deformable stomach model is stationary when no external periodic forces are applied.
\item Dynamic environment (DE):
As described in Section~\ref{method:env}, periodic sinusoidal forces were applied to the deformable stomach model to simulate respiration and natural peristaltic movements. This introduced dynamic deformations in the environment, independent of direct contact with the robotic endoscope. As the target moves, the endoscope must adapt to the constantly changing geometry of the stomach, similar to realistic conditions.
\end{itemize}
\subsubsection{Training and simulation parameters}
The simulation and training environments were implemented using a custom Gym-compatible interface \cite{towers2024gymnasium}, leveraging the deformable behavior modeled by FEM in the SOFA framework \cite{sofa}. The number of maximum steps of training episodes is set to 128. A continuous action space with five degrees of freedom is constrained in the range of [-0.4, 0.4]. The state space consisted of 6 features: the position vector from the robot's tip to the target, the velocity of the robot tip, cable length displacement, prior axial actions, binary contact indicators, and normalized contact forces. Without losing realistic considerations, contact forces can be obtained in real FRE, as demonstrated in soft robotic sensing and contact estimation \cite{nc2024senseexpectation,SRrealcontactsensing1,SRrealcontactsensing2}.

As described in Eq. \ref{Equ:periodicforce}, the dynamics of stomach was simulated by a LinearForceAlternator controller. This alternation induced continuous deformations, requiring the FRE to adapt its trajectory in real-time. A random selection from the predefined list of indices, $IndexList_A$, on the stomach surface defined the variation in navigation goals. Episodes terminated when the distance between the robotic tip and the target was less than 3 mm, the tip of the endoscope was outside of the stomach's boundaries, or the maximum number of steps was exceeded. The network consisted of four layers with sizes [256, 128, 64, 32]. DRL was trained on a 12th Gen Intel® Core™ i7-12700K CPU with 20 cores, the training rate was maintained at 4 Hz.

\subsubsection{Policies}
\label{sec:policydescription}
To evaluate the importance of force-informed observation and dynamic environment, the training process was performed under five settings:

\begin{itemize}
  \item Policy A ($\pi_A$): trained in \textbf{FE}, \textbf{without any contact}.
  \item Policy B ($\pi_B$): trained in the \textbf{SE}, but \textbf{without access to force and contact information} \cite{ng2024navigation}.
  \item Policy C ($\pi_C$): trained in the \textbf{SE}, \textbf{with state information of the resultant force and the contact indicator}.
  \item Policy D ($\pi_D$): trained in the \textbf{DE}, but \textbf{without access to force and contact information}.
  \item Policy E ($\pi_E$): trained in the \textbf{DE}, \textbf{with state information of the resultant force and contact indicator} (our CAN strategy for learning in a challenging environment with the benefit of force awareness).
\end{itemize}
\subsection{Training performances}
\begin{figure}[htbp]
  \centering
  \includegraphics[scale=0.1]{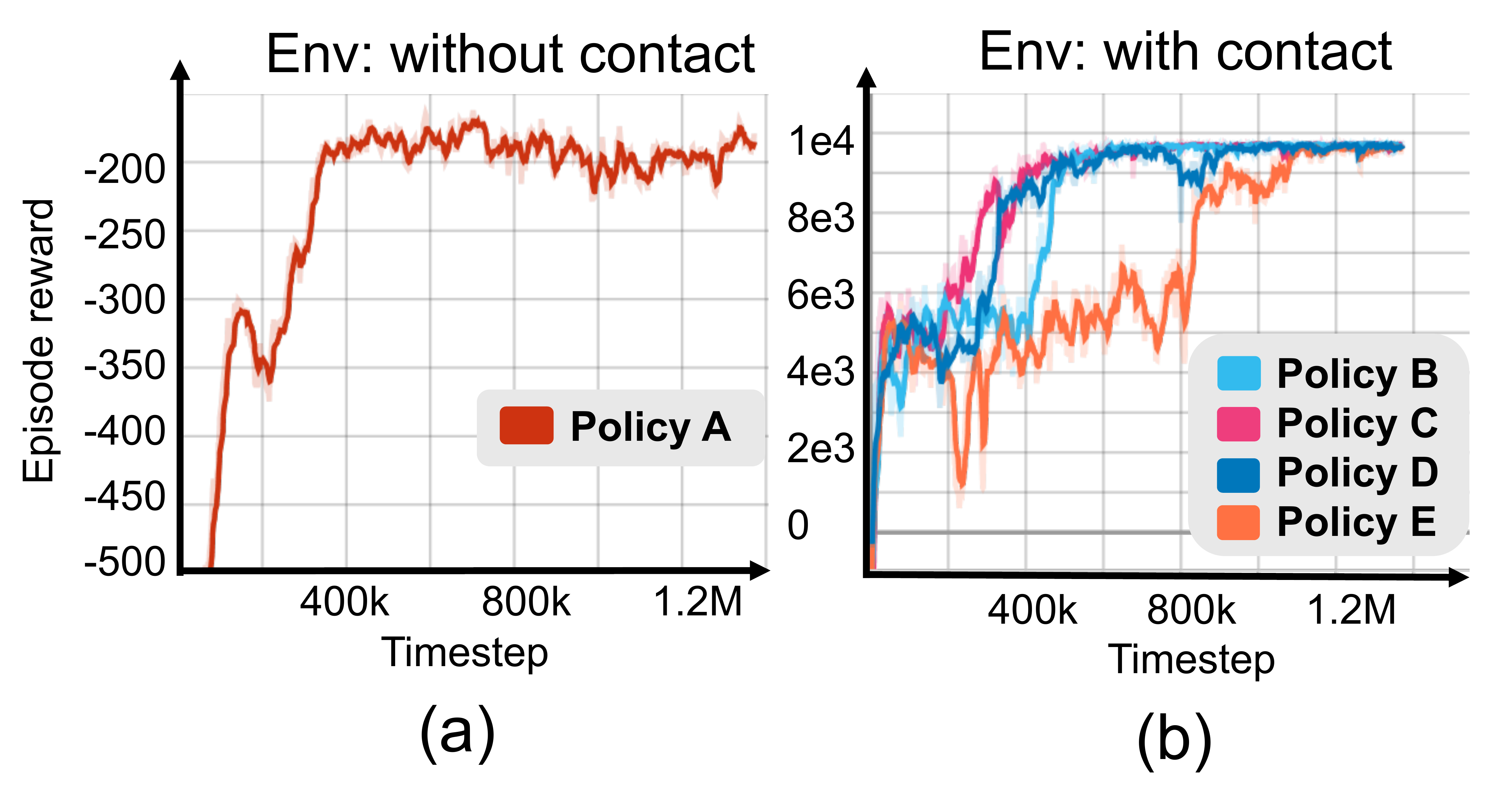}
  \caption{DRL training curves. (a) $\pi_A$ in environment without contact; (b) $\pi_B$, $\pi_C$, $\pi_D$, $\pi_E$ in environments with contact. Their state representations are illustrated in \ref{sec:policydescription}.}
  \label{Fig:trainingcurve}
\end{figure}

The training results of the five policies demonstrate their convergence at varying levels of environmental complexity and input state spaces. As shown in Fig. \ref{Fig:trainingcurve}(a), $\pi_A$ trained in FE, exhibited the fastest convergence around 300k timesteps. The reward of $\pi_A$ is shown in Eq. 6. This indicates that the FE posed minimal challenges, allowing the policy to learn an effective navigation strategy with fewer interactions. 

In contrast, the training curves of $\pi_B,\pi_C,\pi_D,\pi_E $ in Fig. \ref{Fig:trainingcurve}(b), reveal a more gradual convergence. Remarkably, policy E trained with force feedback in state and the most complex environment, converging after 1M timesteps. 

Despite these differences, all the policies in Fig. \ref{Fig:trainingcurve}(b) ultimately converged to similar reward levels within their respective training environments. This indicates that DRL effectively optimized navigation strategies under a consistent reward structure, suggesting that the policies acquired comparable navigation capabilities in their respective settings. As shown in Eq. (6)-(8), under the same reward function, all policies learned to optimize their paths, efficiently explored the stomach region, and maintained a stable success rate (SR) in different target locations. Each policy converged to an episode reward of approximately $10^4$, demonstrating the robustness of DRL in the navigation task. This outcome highlights that, despite variations in environmental complexity and state representations, the trained policies achieved similar performance within their training conditions.

\subsection{Navigation performance}
This section evaluates the navigation performances. The experiments assess navigation in training environments (TE). Then, policies with \textbf{zero-shot} are evaluated in UE with different external force dynamics and unseen target sets. The SR and average error (AE) are used as metrics to compare the performance of the policies. By analyzing the results, we aim to demonstrate the impact of incorporating force and contact feedback into the state representation, particularly in dynamic and complex navigation scenarios (maximum 80 timesteps for each trial).
\subsubsection{Navigation in TE}
$\pi_A$ trained without environmental contact or force feedback, achieved a high SR of 98.3\% in free space, with an impressive average error of 0.4 mm. This indicates that $\pi_A$ is highly effective in simple navigation tasks without contact, where no external forces or environmental interactions are involved. 

\setlength\tabcolsep{0.2em}
\begin{table}[H]
\caption{\textbf{Comparative study of policies SE vs. DE}} 
\centering
\begin{tabular}{@{}lcccc}
\toprule
\textbf{Policy}        & \textbf{SR in SE [\%] $\uparrow$} & \textbf{AE [mm] $\downarrow$} & \textbf{SR in DE [\%] $\uparrow$} & \textbf{AE [mm] $\downarrow$} \\ \midrule
A                      & 26.2                   & 2.5              & 34.5                   & 2.2              \\
B                      & 98                     & 1.6              & 81.5                   & 1.7              \\
C                      & \textbf{100}                    & 1.6              & \textbf{100}                  & 1.6              \\
D                      & 91.8                   & 1.6              & 97.8                   & 1.6              \\
E                      & \textbf{100}                   & 1.6              & \textbf{100}                    & 1.6              \\ 
\bottomrule
\end{tabular}
\label{tab:sr_ae_metrics}
\end{table}

As shown in Table~\ref{tab:sr_ae_metrics}, in the SE, $\pi_C$, and $\pi_E$ achieved a 100\% SR and an AE of 1.6 mm, demonstrating their superior precision and reliability. $\pi_B$ trained in the SE, achieved a slightly lower SR of 98\%. $\pi_A$ trained in FE, performed poorly, with an SR of only 26.2\% and an AE of 2.5 mm. The lack of contact and force feedback when training $\pi_A$ limits its adaptability to complex, dynamic, and constrained environments, which highlights the importance of training in contact environments for improving navigation performance.

In the DE, $\pi_C$, $\pi_E$ maintained a 100\% SR and 1.6 mm AE, adapting effectively to the complex dynamic environment. In contrast, $\pi_D$ showed a slightly lower SR of 97.8\%, and $\pi_B$ achieved only 81.5\% SR, indicating reduced adaptability in dynamic scenarios without access to force and contact data. $\pi_A$'s performance was limited to an SR of 34.5\%, reinforcing the importance of training in a contact environment and adding contact force feedback to state information.

\subsubsection{Navigation in UE}
To evaluate the policies, we construct unstructured environment 1 (UE1) and unstructured environment 2 (UE2) that apply doubled and triple external periodic forces respectively. UE1 and UE2 use unseen target set $IndexList_B$. As shown in Table~\ref{tab:sr_ae_metrics_ue}, $\pi_E$ outperformed all others with SR of 84.5\% in UE1. $\pi_C$ also demonstrated strong performance with an SR of 81\%, but $\pi_B$ and $\pi_D$, trained without force feedback, achieved lower SRs of 75.6\% and 71.3\%. In UE2, with triple periodic forces, the best policy $\pi_E$ maintained the highest SR of 85\%. $\pi_A$ showed the weakest adaptability to UE, achieving around 30\% SR.

\setlength\tabcolsep{0.2em}
\begin{table}[H]
\caption{\textbf{Comparative study of policies UE1 vs. UE2}} 
\centering
\begin{tabular}{@{}lcccc}
\toprule
\textbf{Policy}        & \textbf{SR in UE1 [\%] $\uparrow$} & \textbf{AE [mm] $\downarrow$} & \textbf{SR in UE2 [\%] $\uparrow$} & \textbf{AE [mm] $\downarrow$} \\ \midrule
A                      & 32.5                   & 2.2              & 33.8                   & 2.5              \\
B                      & 75.6                   & 1.8              & 68.8                   & 2.1              \\
C                      & 81                     & 1.9              & 78                     & 1.8              \\
D                      & 71.3                   & 2.1              & 55                     & 2.0              \\
E                      & \textbf{84.5}                 & 1.9              & \textbf{85}                   & 1.9              \\ 
\bottomrule
\end{tabular}
\label{tab:sr_ae_metrics_ue}
\end{table}

Policies trained with this information demonstrated higher SRs and precision in UE, which emphasize the critical role of force and contact feedback in enabling policies to adapt to complex and dynamic environments. 

\begin{figure}[t!]
  \centering
  \includegraphics[scale=0.11]{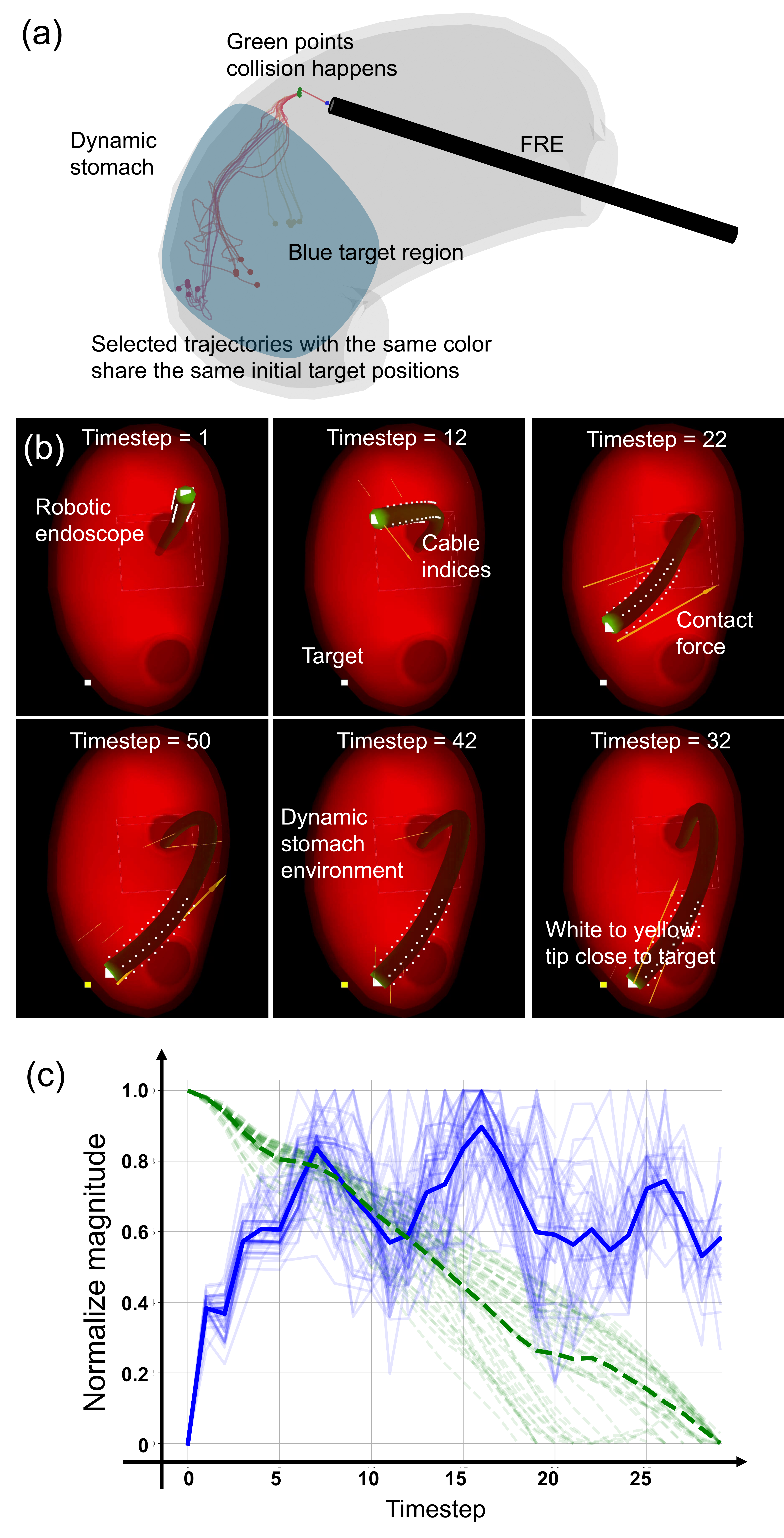}
  \caption{(a) Selected trajectories with the same color share the same initial target positions in the dynamic environment. Targets were randomly selected in the blue region. Green points indicate FRE starts colliding with the stomach. 
  (b) Navigation demonstration by pre-trained $\pi_E$. Yellow arrows are the reaction force between the endoscope and the stomach wall. The white point is an assigned target, which turns yellow if the distance between the EE and the target is less than 3 mm. (c) By pre-trained policy $\pi_E$, randomly select targets to perform navigation tasks. Normalized force magnitude and distance to the target across navigation trajectories, illustrating an increase in force magnitude as the robot tip approaches the target.}
  \label{Fig:demonstraction}
\end{figure}

\subsubsection{RFE navigation demonstration}

Fig. \ref{Fig:demonstraction}(a) presents selected navigation trajectories of the FRE in a dynamic stomach. The trajectories are color-coded, where paths sharing the same color correspond to the same initial target positions. Targets are randomly assigned within the designated blue region. As the robot moves toward a target, it continuously adapts to the changing environment while interacting with the deformable stomach walls. Green points along the trajectories indicate the FRE starts to come into contact with the stomach. The varying trajectory patterns reflect the FRE’s ability to respond dynamically to environmental deformations and external disturbances.

Fig. \ref{Fig:demonstraction}(b) illustrates the navigation process using pre-trained $\pi_E$. The robot actively maneuvers toward the target, represented by a white point, which transitions to yellow once the EE reaches within 3 mm of the target position. Yellow arrows indicate the reaction forces exerted between the endoscope and the stomach wall, providing insight into how contact forces influence navigation. The trajectories demonstrate CAN strategy that force information is needed in DE. These forces arise due to the interaction between FRE and the stomach. By utilizing DRL, the policy enables the robot to adjust its motion based on the force information.

Fig. \ref{Fig:demonstraction}(c) shows a normalized force magnitude and remaining distance between EE and targets across navigation trajectories. The transparent blue and green lines represent individual trajectory data for normalized force magnitude and distance, respectively. The solid blue line indicates the smoothed average of normalized force magnitude across all trajectories, obtained via cubic spline interpolation, while the dashed green line represents the average normalized distance to the target. The data highlights the trend of increasing force magnitude as the robot tip approaches the target, showcasing the robot's reliance on contact-based guidance. The robot tip is far from the target, and the force magnitude is minimal. As the robot moves forward, the contact area between the stomach wall and the passive end of the robot increases, leading to a rise in force magnitude. This trend highlights the progressive interaction between the robot and its environment, with force values intensifying as the robot approaches the target, which shows a clear inverse relationship: the closer the tip gets to the target, the greater the force experienced. This behavior underscores the significance of force information as a navigation signal in FRE navigation.

\section{CONCLUSIONS}
We proposed the CAN strategy to navigate an FRE in dynamic stomach. To address the challenges of non-intuitive manual control, a force-informed DRL framework was developed to enhance autonomy. Contact forces serve as critical guidance signals, allowing the FRE to perform complex movements and improve navigation performance in UE. The effectiveness of force feedback in overcoming the uncertainties in deformable stomachs for FRE is shown.

Despite its advantages, the proposed DRL-based CAN method has limitations. The longer convergence times in complex environments highlight the increased computational cost and learning effort required for policy adaptation. Additionally, while contact forces improve navigation, visual information is also an important observation for navigation in real FRE. Combining information of contact and vision remains an area for investigation. Future research will focus on extending the current state-based DRL policy to a multi-modal DRL framework that integrates both force and vision information. This enhancement will enable the FRE to leverage complementary sensory inputs for more robust navigation. Additionally, investigating safe navigation algorithms is essential to minimize the risk of tissue injury, ensuring that CAN remains both effective and non-invasive.

\balance

\bibliographystyle{ieeetr}
\bibliography{root}

\begin{thebibliography}{10}

\bibitem{chiu2024EndoMaster}
P.~W. Chiu, H.~C. Yip, S.~Chu, S.~M. Chan, H.~S.~L. Lau, R.~S. Tang, S.~J. Phee, K.~Y. Ho, and S.~S.~M. Ng, ``Prospective single-arm trial on feasibility and safety of an endoscopic robotic system for colonic endoscopic submucosal dissection,'' {\em Endoscopy}, 2024.

\bibitem{gaoijrr}
H.~Gao, X.~Yang, X.~Xiao, X.~Zhu, T.~Zhang, C.~Hou, H.~Liu, M.~Q.-H. Meng, L.~Sun, X.~Zuo, {\em et~al.}, ``Transendoscopic flexible parallel continuum robotic mechanism for bimanual endoscopic submucosal dissection,'' {\em The International Journal of Robotics Research}, vol.~43, no.~3, pp.~281--304, 2024.

\bibitem{laitii}
J.~Lai, T.-A. Ren, W.~Yue, S.~Su, J.~Y. Chan, and H.~Ren, ``Sim-to-real transfer of soft robotic navigation strategies that learns from the virtual eye-in-hand vision,'' {\em IEEE Transactions on Industrial Informatics}, 2023.

\bibitem{crreviewTRO}
J.~Burgner-Kahrs, D.~C. Rucker, and H.~Choset, ``Continuum robots for medical applications: A survey,'' {\em IEEE Transactions on Robotics}, vol.~31, no.~6, pp.~1261--1280, 2015.

\bibitem{stosize}
A.~M. Karnul, C.~K. Murthy, and C.~K. Murthy, ``A study of variations of the stomach in adults and growth of the fetal stomach,'' {\em Cureus}, vol.~14, no.~8, 2022.

\bibitem{zhang2019motion}
Z.~Zhang, J.~Dequidt, J.~Back, H.~Liu, and C.~Duriez, ``Motion control of cable-driven continuum catheter robot through contacts,'' {\em IEEE Robotics and Automation Letters}, vol.~4, no.~2, pp.~1852--1859, 2019.

\bibitem{softcontactnav2}
P.~Rao, O.~Salzman, and J.~Burgner-Kahrs, ``Towards contact-aided motion planning for tendon-driven continuum robots,'' {\em IEEE Robotics and Automation Letters}, 2024.

\bibitem{softcontactnav3}
M.~Selvaggio, L.~Ramirez, N.~D. Naclerio, B.~Siciliano, and E.~W. Hawkes, ``An obstacle-interaction planning method for navigation of actuated vine robots,'' in {\em 2020 IEEE International Conference on Robotics and Automation (ICRA)}, pp.~3227--3233, IEEE, 2020.

\bibitem{nc2024senseexpectation}
P.~Wang, Z.~Xie, W.~Xin, Z.~Tang, X.~Yang, M.~Mohanakrishnan, S.~Guo, and C.~Laschi, ``Sensing expectation enables simultaneous proprioception and contact detection in an intelligent soft continuum robot,'' {\em Nature Communications}, vol.~15, no.~1, p.~9978, 2024.

\bibitem{softcontactnav1}
J.~D. Greer, L.~H. Blumenschein, R.~Alterovitz, E.~W. Hawkes, and A.~M. Okamura, ``Robust navigation of a soft growing robot by exploiting contact with the environment,'' {\em The International Journal of Robotics Research}, vol.~39, no.~14, pp.~1724--1738, 2020.

\bibitem{haggerty2019characterizing}
D.~A. Haggerty, N.~D. Naclerio, and E.~W. Hawkes, ``Characterizing environmental interactions for soft growing robots,'' in {\em 2019 IEEE/RSJ International Conference on Intelligent Robots and Systems (IROS)}, pp.~3335--3342, IEEE, 2019.

\bibitem{ng2024navigation}
C.~Ng, H.~Gao, T.-A. Ren, J.~Lai, and H.~Ren, ``Navigation of tendon-driven flexible robotic endoscope through deep reinforcement learning,'' in {\em 2024 IEEE International Conference on Advanced Robotics and Its Social Impacts (ARSO)}, pp.~134--139, IEEE, 2024.

\bibitem{softdrlsimreal}
U.~Berdica, M.~Jackson, N.~E. Veronese, J.~Foerster, and P.~Maiolino, ``Reinforcement learning controllers for soft robots using learned environments,'' in {\em 2024 IEEE 7th International Conference on Soft Robotics (RoboSoft)}, pp.~933--939, IEEE, 2024.

\bibitem{softdrlsofagym}
P.~Schegg, E.~M{\'e}nager, E.~Khairallah, D.~Marchal, J.~Dequidt, P.~Preux, and C.~Duriez, ``Sofagym: An open platform for reinforcement learning based on soft robot simulations,'' {\em Soft Robotics}, vol.~10, no.~2, pp.~410--430, 2023.

\bibitem{softdrldr}
G.~Tiboni, A.~Protopapa, T.~Tommasi, and G.~Averta, ``Domain randomization for robust, affordable and effective closed-loop control of soft robots,'' in {\em 2023 IEEE/RSJ International Conference on Intelligent Robots and Systems (IROS)}, pp.~612--619, IEEE, 2023.

\bibitem{(JIRS23)RL_CR_control}
T.~C. Kargin and J.~Ko{\l}ota, ``A reinforcement learning approach for continuum robot control,'' {\em Journal of Intelligent \& Robotic Systems}, vol.~109, no.~4, p.~77, 2023.

\bibitem{nce24Axisspace}
D.~Wei, J.~Zhou, Y.~Zhu, J.~Ma, and S.~Ma, ``Axis-space framework for cable-driven soft continuum robot control via reinforcement learning,'' {\em Communications Engineering}, vol.~2, no.~1, p.~61, 2023.

\bibitem{elastica}
N.~Naughton, J.~Sun, A.~Tekinalp, T.~Parthasarathy, G.~Chowdhary, and M.~Gazzola, ``Elastica: A compliant mechanics environment for soft robotic control,'' {\em IEEE Robotics and Automation Letters}, vol.~6, no.~2, pp.~3389--3396, 2021.

\bibitem{gmsh}
C.~Geuzaine and J.-F. Remacle, ``Gmsh: A 3-d finite element mesh generator with built-in pre-and post-processing facilities,'' {\em International Journal for Numerical Methods in Engineering}, vol.~79, no.~11, pp.~1309--1331, 2009.

\bibitem{sofa}
F.~Faure, C.~Duriez, H.~Delingette, J.~Allard, B.~Gilles, S.~Marchesseau, H.~Talbot, H.~Courtecuisse, G.~Bousquet, I.~Peterlik, {\em et~al.}, ``{SOFA}: A multi-model framework for interactive physical simulation,'' {\em Soft Tissue Biomechanical Modeling for Computer Assisted Surgery}, pp.~283--321, 2012.

\bibitem{towers2024gymnasium}
M.~Towers, A.~Kwiatkowski, J.~Terry, J.~U. Balis, G.~De~Cola, T.~Deleu, M.~Goul{\~a}o, A.~Kallinteris, M.~Krimmel, A.~KG, {\em et~al.}, ``Gymnasium: A standard interface for reinforcement learning environments,'' {\em arXiv preprint arXiv:2407.17032}, 2024.

\bibitem{SB3}
A.~Raffin, A.~Hill, A.~Gleave, A.~Kanervisto, M.~Ernestus, and N.~Dormann, ``Stable-baselines3: Reliable reinforcement learning implementations,'' {\em Journal of Machine Learning Research}, vol.~22, no.~268, pp.~1--8, 2021.

\bibitem{trainlearn}
J.~Ibarz, J.~Tan, C.~Finn, M.~Kalakrishnan, P.~Pastor, and S.~Levine, ``How to train your robot with deep reinforcement learning: lessons we have learned,'' {\em The International Journal of Robotics Research}, vol.~40, no.~4-5, pp.~698--721, 2021.

\bibitem{PPO}
J.~Schulman, F.~Wolski, P.~Dhariwal, A.~Radford, and O.~Klimov, ``Proximal policy optimization algorithms,'' {\em arXiv preprint arXiv:1707.06347}, 2017.

\bibitem{SRrealcontactsensing1}
T.~G. Thuruthel, B.~Shih, C.~Laschi, and M.~T. Tolley, ``Soft robot perception using embedded soft sensors and recurrent neural networks,'' {\em Science Robotics}, vol.~4, no.~26, p.~eaav1488, 2019.

\bibitem{SRrealcontactsensing2}
A.~Gao, Z.~Lin, C.~Zhou, X.~Ai, B.~Huang, W.~Chen, and G.-Z. Yang, ``Body contact estimation of continuum robots with tension-profile sensing of actuation fibers,'' {\em IEEE Transactions on Robotics}, vol.~40, pp.~1492--1508, 2024.

\end{thebibliography}
\end{document}